\documentclass{article}
\usepackage{amsmath}
\usepackage{bm} 
\usepackage{graphicx}
\usepackage{caption}
\usepackage{algorithm}
\usepackage{algpseudocode}
\usepackage{amsfonts} 

\PassOptionsToPackage{numbers, compress}{natbib}
\usepackage[dblblindworkshop, final]{neurips_2025}
\workshoptitle{AI for Science}
\makeatletter
\renewcommand{\@trackname}{NeurIPS 2025 AI for Science Workshop}
\makeatother

\usepackage[utf8]{inputenc} 
\usepackage[T1]{fontenc}    
\usepackage{hyperref}       
\usepackage{url}            
\usepackage{booktabs}       
\usepackage{amsfonts}       
\usepackage{nicefrac}       
\usepackage{microtype}      
\usepackage{xcolor}         

\title{AIM: Adaptive Intervention for Deep Multi-task Learning of Molecular Properties}

\author{%
   Mason Minot, Gisbert Schneider \\
   Department of Biosystems Science and Engineering\\
   ETH Z{\"u}rich\\
   Basel, 4056, Switzerland  \\
   \texttt{mminot@ethz.ch, gisbert.schneider@bsse.ethz.ch}
}

\begin{document}

\maketitle

\begin{abstract}
Simultaneously optimizing multiple, frequently conflicting, molecular properties is a key bottleneck in the development of novel therapeutics. Although a promising approach, the efficacy of multi-task learning is often compromised by destructive gradient interference, especially in the data-scarce regimes common to drug discovery. To address this, we propose AIM, an optimization framework that learns a dynamic policy to mediate gradient conflicts. The policy is trained jointly with the main network using a novel augmented objective composed of dense, differentiable regularizers. This objective guides the policy to produce updates that are geometrically stable and dynamically efficient, prioritizing progress on the most challenging tasks. We demonstrate that AIM achieves statistically significant improvements over multi-task baselines on subsets of the QM9 and targeted protein degraders benchmarks, with its advantage being most pronounced in data-scarce regimes. Beyond performance, AIM's key contribution is its interpretability; the learned policy matrix serves as a diagnostic tool for analyzing inter-task relationships. This combination of data-efficient performance and diagnostic insight highlights the potential of adaptive optimizers to accelerate scientific discovery by creating more robust and insightful models for multi-property molecular design.
\end{abstract}

\section{Introduction}
The \textit{in silico} design of molecules frequently requires the simultaneous optimization of multiple, often competing, objectives~\cite{schneider2020rethinking,atz2024geometric}. For instance, a candidate drug molecule must exhibit high binding affinity to its target while also possessing favorable pharmacokinetic properties such as aqueous solubility and metabolic stability~\cite{nippa2025expediting}. In domains such as drug discovery, where high-quality experimental data is often scarce and costly to acquire, achieving data efficiency is crucial~\citep{anelli2024robust, minot2024meta, komissarov2024actionable, minot2025active}. Multi-task learning (MTL) is a strategy to this end, as it enables a single model to learn a unified representation across related tasks~\cite{allenspach2024neural}.

However, the efficacy of MTL is often compromised by negative transfer~\cite{liu2021conflict, yu2020gradient}. When task objectives conflict, their gradients can pull the model in opposing directions, leading to destructive interference that degrades overall performance. As a result, MTL's potential for data efficiency is often unrealized because the very tasks meant to help one another end up in conflict. To effectively leverage MTL, it is therefore desirable to minimize negative transfer while maximizing synergistic benefits. 

A considerable body of work has sought to mitigate this negative interference, laying a crucial foundation for robust multi-task optimization. Methods such as PCGrad~\cite{yu2020gradient}, CAGrad~\cite{liu2021conflict}, and Nash-MTL~\cite{navon2022multi} mitigate conflicts using principled, but often static, rules. Although effective, the uniform strategy applied by these hand-crafted heuristics creates an opportunity for a learned, adaptive policy to discover more nuanced strategies for navigating these complex trade-offs. \\

In this work, we reframe the problem from one of heuristic gradient surgery to one of learned, adaptive optimization. We propose Adaptive Intervention for deep Multi-task learning of molecular properties (AIM), an optimization framework that learns a dynamic, context-aware policy to transform conflicting gradients into a more effective update step. The policy is trained jointly with the main model by optimizing a novel augmented objective guided by a generalization signal from a held-out validation set and is augmented with dense, differentiable regularizers. Key regularizers ensure dynamic efficiency by preserving the update step's magnitude and promoting progress on tasks with the highest current loss. \\

We validate AIM on a 2D toy problem, the foundational QM9 dataset~\cite{ramakrishnan2014quantum}, and a complex targeted protein degrader (TPD) ADME (absorption, distribution, metabolism, and excretion) benchmark~\cite{peteani2024application}. By evaluating on data subsets to simulate real-world, data-scarce scenarios, we show that AIM's learned policy discovers more effective optimization strategies than handcrafted heuristics, particularly in low-data regimes.

\section{Related work} 
This section lays the groundwork for our proposed method. We begin by formally introducing the MTL paradigm and its central challenge of destructive interference. We then review the foundational heuristic approaches, setting the stage for the adaptive learning-based methods we explore. Finally, we introduce the concept of a learned optimization policy as a natural extension of this prior work.

\subsection{The challenge of multi-task optimization}

Training a single neural network for multiple scientific tasks holds the promise of improving data efficiency and discovering shared representations between physical properties. However, this MTL paradigm introduces a fundamental optimization challenge, namely navigating the trade-offs between competing task objectives.

\subsection{Heuristics for mitigating gradient interference}
The common formulation of the MTL objective is to minimize a weighted sum of individual task losses $\mathcal{L}(\theta)=\sum_{i}w_i\mathcal{L}_i(\theta)$. A primary obstacle is destructive gradient interference, where gradients from different tasks $\mathbf{g_i} = \nabla_\theta \mathcal{L}_i$, pull model parameters $\theta$ in competing directions. This interference is multifaceted:

\begin{enumerate}
    \item \textbf{Directional conflict:} The literature has traditionally defined directional conflict as an instance where gradients are opposed ($\mathbf{g_i} \cdot \mathbf{g_j} < 0$)~\cite{yu2020gradient}. Building on this, our work proposes a more flexible definition, positing that conflict exists across a spectrum of misalignments. Even positively correlated gradients can represent a suboptimal compromise, as their vector sum may neglect the needs of other tasks. The core challenge is thus not simply to avoid opposed gradients, but to learn the optimal degree of alignment for each specific pair of tasks.
    \item \textbf{Magnitude conflict:} A task with a large gradient norm can dominate the update, hindering progress on other tasks. Although often treated separately, magnitude and directional conflicts are intertwined; an imbalance in magnitude can exacerbate directional disputes. An effective intervention mechanism, therefore, should ideally account for both.
\end{enumerate}

The presence of task conflicts means that a single optimal solution is rarely feasible, leading to a Pareto front of trade-off solutions. The difficulty lies in choosing the point on this front that offers the best generalization. Whereas static heuristics address this with a fixed geometric rule, our work learns this selection process directly from the data, enabling an adaptive strategy.

 \subsection{Navigating the front: from heuristics to a data-driven approach} A variety of methods have been developed to navigate the multi-task optimization landscape. The simplest approach, linear scalarization (LS), which minimizes the aggregate loss, often fails to converge to the Pareto front. Principled geometric and game-theoretic methods, including PCGrad~\cite{yu2020gradient}, CAGrad~\cite{liu2021conflict}, and Nash-MTL~\cite{navon2022multi}, apply fixed geometric rules to gradients. Complementary to these, adaptive weighting heuristics like FAMO~\cite{liu2023famo} balance tasks by adjusting their weights based on relative progress.  Our work explores a more flexible paradigm in which the intervention policy itself is learned, enabling an adaptive strategy that is sensitive to the unique relationships between tasks.

\section{Method: Adaptive Intervention for deep Multi-task learning (AIM)}

AIM reframes the resolution of gradient conflicts from a fixed heuristic to a learned, adaptive policy $\Psi$. This policy transforms a set of raw task gradients $\{\mathbf{g}_i\}$ into a unified update vector $\mathbf{g}_{\text{intervened}}$. This transformed gradient is then used to update the main model parameters, $\theta$, in a standard optimization step. The policy's innovation lies in its data-driven definition of gradient conflict, which is trained by optimizing a novel augmented objective.

\subsection{Differentiable gradient intervention}
Instead of relying on a static geometric rule (e.g., intervening only when $\mathbf{g}_i \cdot \mathbf{g}_j < 0$), AIM's policy learns when and how strongly to intervene based on a learnable conflict threshold, $\tau$. Crucially, the policy treats the raw task gradients $\{\mathbf{g}_i\}$ as fixed input vectors. Although formal meta-learning~\cite{finn2017model} or second-order methods could also guide the policy, such approaches can be computationally prohibitive. Instead, by applying a stop-gradient, we prevent the need to differentiate through the gradient computation itself, thus avoiding expensive second-order derivatives while retaining the benefits of a learned, adaptive policy. 

The intervention is performed in a pairwise manner. For each task gradient $\mathbf{g}_i$, its modified counterpart $\mathbf{g}_i'$ is computed by considering its relationship with all other task gradients $\mathbf{g}_j$ (where $j \neq i$). The strength of intervention between any pair is determined by a soft, differentiable projection weight $w_{\text{proj}}^{(i,j)}$:
\begin{equation}
w_{\text{proj}}^{(i,j)} = \sigma\left(k \cdot (\tau_{ij} - \cos(\mathbf{g}_i, \mathbf{g}_j))\right)
\end{equation}
where $\sigma(\cdot)$ is the sigmoid function, $k$ is a temperature parameter, and $\tau_{ij}$ is the learnable threshold for the pair. The modified gradient $\mathbf{g}_i'$ is formed by iteratively removing the conflicting components from other gradients, adapting the formulation from PCGrad~\cite{yu2020gradient}:
\begin{equation}
\mathbf{g}_i' = \mathbf{g}_i - \sum_{j \neq i} w_{\text{proj}}^{(i,j)} \cdot \text{proj}_{\mathbf{g}_j}(\mathbf{g}_i)
\end{equation}
where $\text{proj}_{\mathbf{g}_j}(\mathbf{g}_i) = (\frac{\mathbf{g}_i \cdot \mathbf{g}_j}{\|\mathbf{g}_j\|^2}) \mathbf{g}_j$ is the vector projection of $\mathbf{g}_i$ on $\mathbf{g}_j$. These modified gradients, $\{\mathbf{g}'_i\}$, are then summed to form the final update vector, $\mathbf{g}_{\text{intervened}} = \sum_{i=1}^{N} \mathbf{g}'_i$, which is used to update the model parameters: $\theta_{t+1} \leftarrow \text{Adam}(\theta_t, \mathbf{g}_{\text{intervened}})$. The complete process is detailed in Algorithm \ref{alg:aim_update} in the Appendix.

We explore two policy variants:
\begin{enumerate}
    \item \textbf{Scalar policy:} Learns a single global threshold $\tau$, applying a uniform conflict tolerance across all pairs of tasks.
    \item \textbf{Matrix policy:} Learns a unique threshold $\tau_{ij}$ for each pair of tasks, allowing the model to capture nuanced pairwise relationships and providing an interpretable diagnostic tool.
\end{enumerate}

\subsection{Guiding the policy}
The policy parameters, $\Phi$, are trained jointly with the main model by minimizing an augmented objective, $\mathcal{L}_{\text{policy}}$. To guide the policy with a generalization signal, the training data is partitioned once before training into two disjoint datasets: a primary set for calculating task gradients and a policy guidance set (10\% of the training data) for evaluating the policy. At each training step, a batch from the primary set is used to generate the raw gradients, while a separate batch from the policy guidance set is used to update the policy's parameters. This use of a held-out dataset is crucial, as it prevents the policy from learning a trivial solution (e.g., outputting a zero vector) by forcing it to produce updates that generalize to unseen data.

While a forward-looking objective could evaluate the policy based on a simulated "lookahead" step, we found this approach to be computationally expensive. Instead, we opt for a more efficient joint-training scheme where the policy guidance loss, $\mathcal{L}_{\text{guide}}$, is calculated on the current model parameters. This introduces a slight temporal mismatch, but we found it to be a favorable trade-off for the significant gains in computational efficiency. The objective is thus guided by the model's loss on the held-out data and regularizers that promote dynamic efficiency,  with each component weighted by a scalar hyperparameter $\lambda$:

\begin{equation}
\mathcal{L}_{\text{policy}} = \lambda_g  \mathcal{L}_{\text{guide}} + \lambda_m \mathcal{L}_{\text{magnitude}} + \lambda_p \mathcal{L}_{\text{progress}}
\end{equation}

The components of this objective are:
\begin{itemize}
    \item \textbf{$\mathcal{L}_{\text{guide}}$ (Policy guidance objective):} The model loss on a held-out subset of the training data, providing a direct signal of generalization performance.
    \item \textbf{$\mathcal{L}_{\text{magnitude}}$ (Magnitude preservation):} A stability penalty, formulated as $(\|\mathbf{g}_{\text{intervened}}\| - \sum_i \|\mathbf{g}_i\|)^2$, that encourages the policy to redistribute, not destroy, gradient energy.
    \item \textbf{$\mathcal{L}_{\text{progress}}$ (Progress penalty):} A penalty that rewards interventions for aiding tasks with high current loss. It is formulated as $-\sum_i \alpha_i \cdot p_i$, where the weight $\alpha_i = \mathcal{L}_i / \sum_j \mathcal{L}_j$ is the normalized loss of task $i$, and $p_i = \mathbf{g}_{\text{intervened}} \cdot \mathbf{g}_i$ is the projected progress. This encourages the policy to prioritize helping the most difficult tasks at each step.
\end{itemize}
By learning an intervention policy guided by this multi-faceted objective, AIM develops a nuanced, context-aware strategy that overcomes certain limitations of handcrafted heuristics.

\section{Experiments and results}

To evaluate our proposed optimization framework, we compare its performance against several established MTL baselines. To evaluate the effectiveness of the model-agnostic AIM optimizer, we use a consistent graph neural network (GNN) architecture for all methods to ensure a fair comparison, adopting the standard message-passing model used previously~\cite{liu2023famo}. All performance metrics reported are the mean and standard deviation computed over \textit{N} = 3 independent runs with different random seeds. For all experiments on a given benchmark, the test set was held fixed to ensure a fair comparison across different training subset sizes.

\subsection{Visualizing learning trajectories on a toy problem} 
\begin{figure}[h!]
    \centering
    
    \captionsetup{skip=2pt, font=small} 
    
    \includegraphics[scale=0.3]{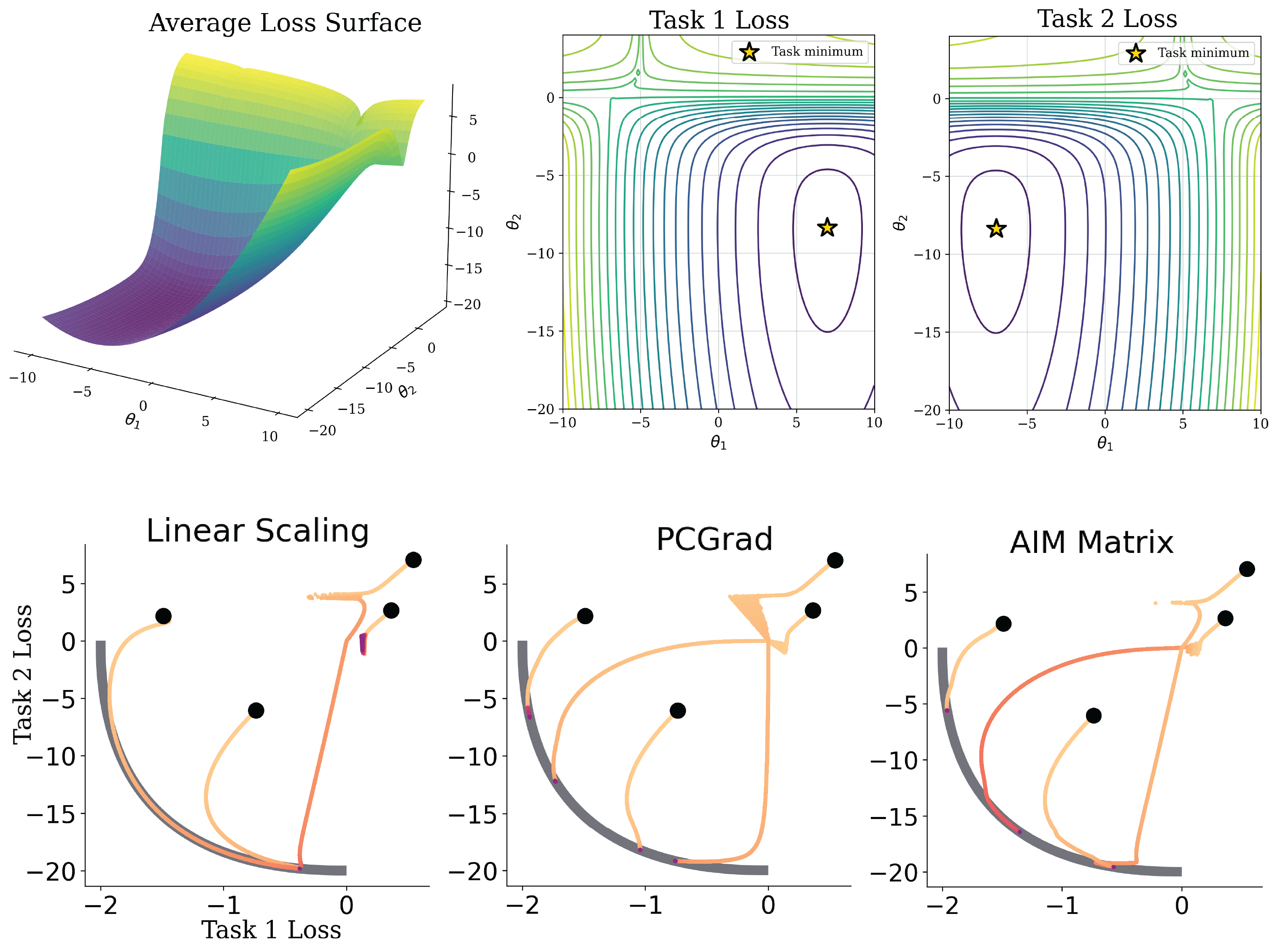}
    
    \caption{(Top) Loss landscape of the 2D toy problem, showing conflicting task minima (stars) and the resulting Pareto front. (Bottom) Optimization trajectories for linear scalarization (LS), PCGrad, and AIM on the conflicted landscape.}
    \label{fig:toy_landscape}
\end{figure}

To visualize how different algorithms navigate a conflicted optimization landscape, we use a two-task problem with a known Pareto front, as shown in Figure \ref{fig:toy_landscape}. We compare the optimization trajectories of LS, the heuristic-based solver PCGrad, and our proposed method, AIM (Matrix). The results highlight a key difference in behavior: LS fails to converge to the Pareto front, whereas both PCGrad and AIM successfully navigate the landscape to find an optimal trade-off solution. This demonstrates that AIM can effectively resolve gradient conflicts, matching the performance of established methods in this idealized setting.

\subsection{QM9 benchmark}

The QM9 dataset~\cite{ramakrishnan2014quantum} serves as a foundational benchmark in computational chemistry, comprising approximately 134,000 small organic molecules and their corresponding quantum mechanical properties, calculated using Density Functional Theory (DFT). These properties, ranging from electronic characteristics such as $\epsilon_{HOMO}$ and $\epsilon_{LUMO}$ to energetic values such as the internal energy at 298.15 K ($U$), provide a diverse and challenging set of prediction targets.

\subsubsection{The learned policy as a diagnostic tool} 
\begin{figure}[h!]
    \centering
    \captionsetup{skip=2pt, font=small} 
    \includegraphics[width=0.9\textwidth]{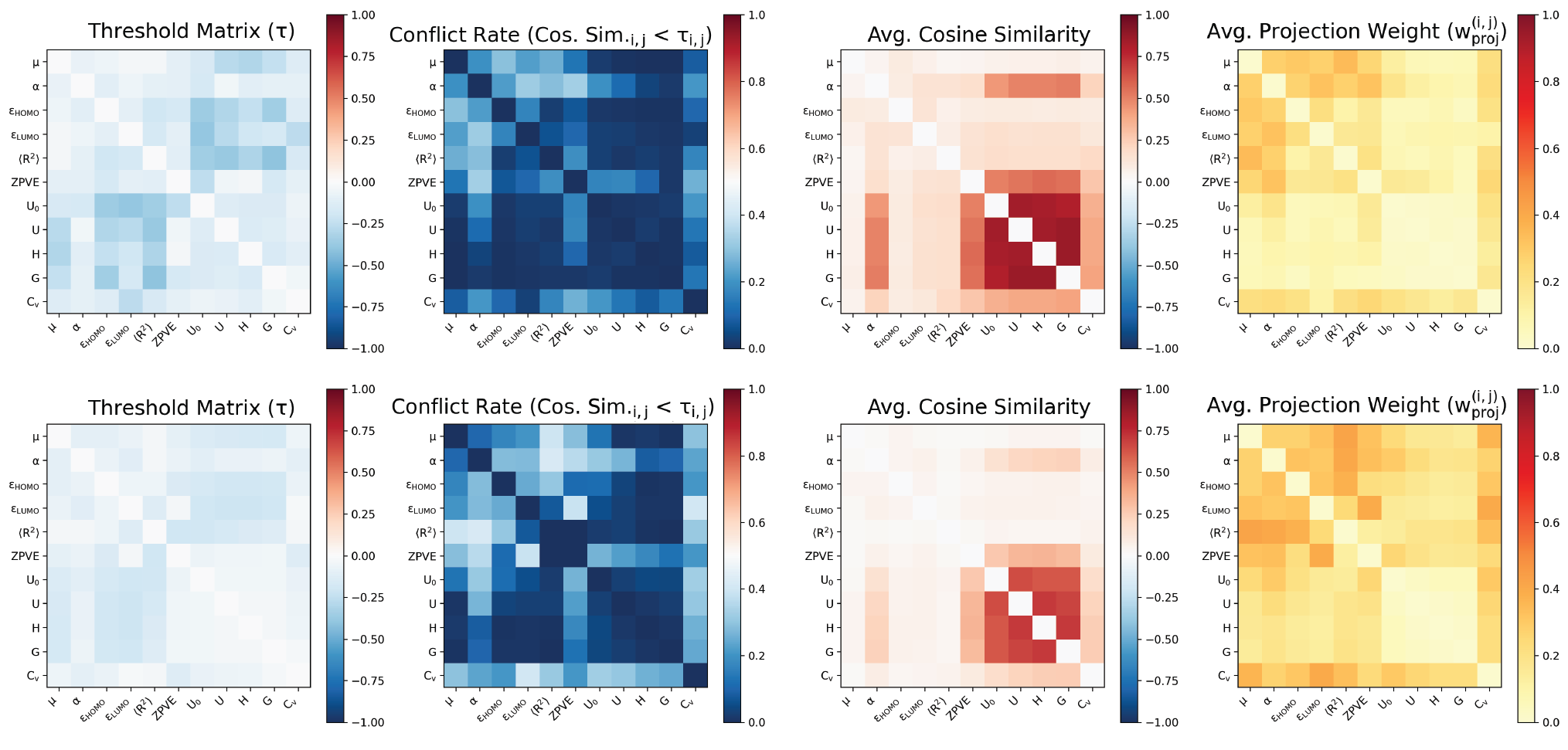} 
    \caption{\textbf{Evolution of the learned AIM policy on QM9.} The heatmaps show four diagnostic metrics for the policy early in training (Epoch 10, top) and later in training (Epoch 100, bottom). From left to right: learned conflict thresholds ($\tau$), conflict rate, raw gradient cosine similarity, and final projection weights.}
    \label{fig:qm9_heatmap}
\end{figure}

A key advantage of AIM is that its learned policy matrix offers a diagnostic window into the MTL process. Figure \ref{fig:qm9_heatmap} visualizes the policy's evolution, showing the learned conflict thresholds, the rates of conflict, raw gradient similarities, and resulting projection weights early (Epoch 10) and later (Epoch 100) in training. The policy learns an intuitive optimization strategy based on the alignment of task gradients, guided by its objective to aid struggling tasks. This is evident with the highly aligned energetic properties ($U_0$, $U$, $H$, $G$), where the policy applies near-zero projection weights. In contrast, as training progresses, stricter interventions for pairs like $C_v$ and $\epsilon_{LUMO}$ drive a more independent optimization. This evolution demonstrates AIM's ability to adapt to nuanced task relationships, a crucial capability for navigating complex optimization landscapes.

\subsubsection{Quantitative performance}

As a strong performance benchmark, we use a single-task learning (STL) baseline, which consists of a separate model trained independently for each task on the full dataset. We evaluate our multi-task learning (MTL) approach against this baseline using Mean Rank (MR) across all tasks and $\Delta_m\%$, the mean percentage error improvement relative to STL. For both metrics, lower values indicate better performance.

The results in Table~\ref{tab:qm9_results} highlight the strength of AIM in low data settings. In the 10k subset, both AIM (Scalar) and AIM (Matrix) significantly outperform all baselines on both MR and $\Delta_m\%$. This trend continues on the 50k subset, where AIM (Scalar) achieves the best performance on both metrics, indicating a robust and balanced performance across the diverse quantum chemical properties. As the dataset size increases to 100k, other methods become more competitive. While FAMO achieves the best $\Delta_m\%$, AIM (Scalar) secures the top Mean Rank, demonstrating that the method remains a strong performer. This overall pattern suggests that AIM's learned policy is particularly effective when data is limited. For a granular view, the MAE per task on the 100k subset is detailed in the Appendix.

\begin{table}[h!]
\centering
\caption{Benchmark results on the QM9 dataset for different training subset sizes. We report the test delta ($\Delta_m\%$) and the mean rank (MR) across all tasks. Lower values are better.}
\label{tab:qm9_results}
\begin{tabular}{lcccccc}
\toprule
& \multicolumn{2}{c}{10k} & \multicolumn{2}{c}{50k} & \multicolumn{2}{c}{100k} \\
\cmidrule(lr){2-3} \cmidrule(lr){4-5} \cmidrule(lr){6-7}
Model & $\Delta_m\%$ & MR & $\Delta_m\%$ & MR & $\Delta_m\%$ & MR \\
\midrule
LS & 588.6 ± 50.3 & 10.82 & 276.1 ± 23.9 & 6.36 & $163.6 \pm 9.6$ & 5.00 \\
PCGrad & 512.5 ± 44.2 & 9.18 & 203.0 ± 10.7 & 3.73 & $132.5 \pm 1.8$ & 4.18 \\
NASH-MTL & 608.8 ± 24.1 & 12.18 & 273.0 ± 25.9 & 5.45 & $185.4 \pm 13.8$ & 6.18 \\
FAMO & 559.0 ± 68.5 & 7.09 & 205.9 ± 31.3 & 3.55 & $\bm{100.6 \pm 12.0}$ & 3.55 \\
CAGrad & 514.6 ± 23.3 & 9.18 & 188.6 ± 16.2 & 4.09 & $116.2 \pm 11.5$ & 4.09 \\
\midrule
AIM (Scalar) & 478.9 ± 23.1 & 8.27 & \textbf{186.4 ± 12.4} & \textbf{2.36} & $120.2 \pm 0.7$ & \textbf{2.27} \\
AIM (Matrix) & \textbf{436.6 ± 25.6} & \textbf{4.45} & 194.5 ± 8.6 & 2.45 & $124.7 \pm 2.1$ & 2.73 \\
\bottomrule
\end{tabular}
\end{table}

\subsection{Targeted protein degraders (TPD) ADME benchmark}

We test our method on a frontier drug discovery challenge: predicting ADME properties for TPDs and other molecules using a recent benchmark~\cite{peteani2024application}. The dataset's 274,00 molecules and 25 diverse tasks provide a complex test-bed for MTL.

\begin{figure}[h!]
    \centering
    \includegraphics[width=1.0\textwidth]{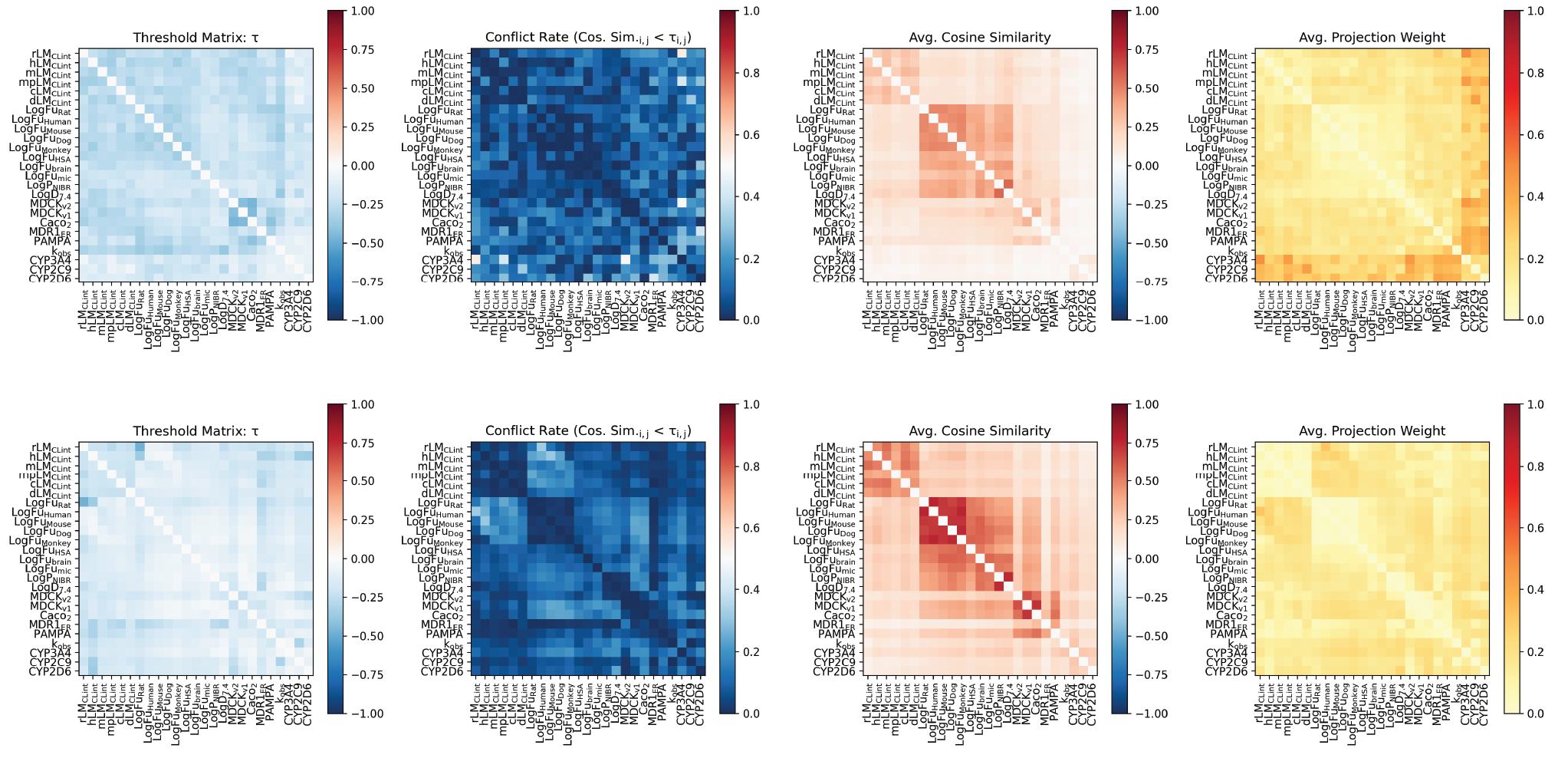} 
    \caption{\textbf{Evolution of the learned AIM policy on TPD.} The heatmaps show the policy's diagnostic metrics early (Epoch 10, top) and late (Epoch 300, bottom) in training. From left to right: learned conflict thresholds ($\tau$), conflict rate, raw gradient similarity, and final projection weights.}
    \label{fig:tpd_heatmap}
\end{figure}

Our evaluation on the TPD ADME dataset demonstrates the effectiveness of our AIM framework. We report the $\Delta_m\%$ over the single-task learning (STL) baseline, which was trained on the full dataset for each respective task, and the MR across all 25 tasks for training data subsets of 10k, 100k and 250k compounds. Similar to the QM9 benchmark, the evolution of the learned policy on the TPD dataset provides a diagnostic view of the inter-task relationships over time (Figure \ref{fig:tpd_heatmap}). Initially, it applies moderate interventions to navigate gradient conflicts and aid tasks like the CYP inhibition assays. This initial corrective action successfully steers the optimization into a more cooperative state later in training, characterized by higher gradient alignment that sharpens and refines the inherent clusters among functionally related tasks.

\begin{table}[h!]
\centering
\caption{Benchmark results on the TPD ADME dataset. We report the test delta ($\Delta_m\%$) and the mean rank (MR) across all tasks. Lower values are better.}
\label{tab:tpd_main_results}
\begin{tabular}{lcccccc}
\toprule
& \multicolumn{2}{c}{10k} & \multicolumn{2}{c}{100k} & \multicolumn{2}{c}{250k} \\
\cmidrule(lr){2-3} \cmidrule(lr){4-5} \cmidrule(lr){6-7}
Model & $\Delta_m\%$ & MR & $\Delta_m\%$ & MR & $\Delta_m\%$ & MR \\
\midrule
LS & $117.5 \pm 15.3$ & 5.40 & $34.7 \pm 6.7$ & 5.68 & $19.4 \pm 7.5$ & 4.40 \\
PCGrad & $310.7 \pm 170.3$ & 6.96 & $66.1 \pm 22.1$ & 6.92 & $29.2 \pm 10.5$ & 4.50 \\
NASH-MTL & $116.3 \pm 4.1$ & 5.04 & $30.5 \pm 5.3$ & 3.52 & $19.0 \pm 5.8$ & 3.76 \\
FAMO & $110.6 \pm 4.5$ & 4.12 & $30.5 \pm 3.7$ & 3.76 & $18.6 \pm 4.1$ & 3.80 \\
CAGrad & $105.0 \pm 9.3$ & 3.28 & $28.2 \pm 1.3$ & 3.12 & $\bm{15.0 \pm 1.7}$ & \textbf{2.56} \\
\midrule
AIM (Scalar) & $90.5 \pm 1.4$ & 1.92 & $\bm{28.0 \pm 1.1}$ & 2.52 & $19.4 \pm 3.3$ & 3.84 \\
AIM (Matrix) & $\bm{88.6 \pm 3.3}$ & \textbf{1.28} & $28.1 \pm 3.4$ & \textbf{2.48} & $16.4 \pm 1.2$ & 2.64 \\
\bottomrule
\end{tabular}
\end{table}

AIM delivers state-of-the-art performance, particularly in the more data-scarce regime (Table~\ref{tab:tpd_main_results}). In the 10k and 100k subsets, the AIM methods outperform all baselines. In the 100k subset, for example, the AIM (Matrix) and AIM (Scalar) variants achieve the best and second-best MR (2.48 and 2.52, respectively) and the second-best and best $\Delta_m\%$ (28.1 and 28.0, respectively). As the dataset size increases to 250k, AIM remains competitive. AIM (Matrix) achieves the second-best performance in both $\Delta_m\%$ (16.4) and Mean Rank (2.64), closely following the top-performing CAGrad. This highlights AIM's ability to find a robust and well-balanced solution across a wide array of competing objectives. 

The detailed results per task on the 100k subset are provided in the Appendix. Together, these results demonstrate AIM's strong performance in specific molecular properties related to absorption, distribution, and metabolism.

\section{Discussion}

Our work introduced AIM, an optimization framework that learns an adaptive policy to mediate gradient conflicts in multi-task learning. The experimental results across a toy problem, the QM9 benchmark, and a frontier drug discovery problem, given by the TPD ADME dataset, confirm our central hypothesis: A learned, data-driven intervention policy can outperform static heuristics, particularly in the data-scarce regimes common to early phases of scientific discovery. Importantly, the learned policy itself serves as an interpretable diagnostic tool that offers insights into the relationships between tasks. This discussion synthesizes these findings, addresses the performance nuances, and frames the practical utility of AIM in light of its demonstrated strengths and limitations.

\subsection{The value and trade-offs of an adaptive policy} A key finding of this work is that the value of AIM's adaptive policy is most pronounced when data is limited. In both the QM9 and TPD ADME benchmarks, AIM's advantage over heuristic-based methods like CAGrad and FAMO was greatest on the smaller data subsets. This performance trend leads us to hypothesize that the primary value of an adaptive policy is in discovering a more data-efficient optimization trajectory. In data-limited settings, the fixed geometric rules of heuristics may be too rigid, whereas AIM's learned policy appears to find a more effective path for navigating gradient conflicts. As the amount of training data increases, this advantage narrows, and simpler heuristics become more competitive. We therefore conclude that AIM's core strength provides the most value when data, not compute, is the primary bottleneck. This complexity-versus-robustness trade-off is also observed within AIM itself. The simpler AIM (Scalar) variant sometimes outperforms the more granular Matrix policy, a result we attribute to the scalar policy's reduced risk of overfitting to the training data, making it a more robust, albeit less nuanced, choice in certain scenarios.

\subsection{From black box to diagnostic: interpreting the learned policy}

Beyond pure performance, AIM's primary contribution is its interpretability. The evolution of the policy matrix validates the learning process, showing how AIM learns a sophisticated optimization strategy that adapts to the natural alignment of task gradients. On QM9 (Figure \ref{fig:qm9_heatmap}), for instance, the policy preserves beneficial gradient sharing between highly-aligned energetic properties ($U_0, U, H, G$) with minimal intervention, while enforcing independent optimization paths for non-aligned tasks through stricter interventions. This diagnostic capability extends to the more complex TPD benchmark (Figure \ref{fig:tpd_heatmap}), where the policy's evolution reveals a progression from applying moderate interventions early in training to fostering a more cooperative state characterized by refined task clusters and higher gradient alignment. The ability to learn and visualize these nuanced strategies, which align with established physical and ADME principles, builds confidence in the model and its predictions.

\subsection{The practical utility of multi-task learning in scientific discovery}
A crucial observation is that the single-task learning (STL) baseline sets a formidable performance benchmark. It is important to highlight that the STL baseline was trained on the full dataset for each task, whereas the MTL models had access to only a fraction of the data. Although this data disparity explains much of the performance gap in low-data regimes, a gap persists even when MTL models are trained on larger subsets, particularly for QM9. This points to a fundamental trade-off: an STL model is a specialist, dedicating its entire capacity to the nuances of a single task. An MTL model, in contrast, must find a single, generalized representation that serves as a compromise across multiple, often conflicting, objectives. As a result, for some datasets, including QM9, it has become common practice in the literature to train a single model per task.

However, this idealized STL scenario is often impractical. In scientific discovery, particularly in fields like drug development, the simultaneous optimization of many properties is a hard constraint. Training and deploying hundreds of specialist models is logistically and computationally infeasible, making multi-task learning (MTL) the only viable path forward for creating a single, unified, and operationally efficient model. The fundamental challenge of MTL, therefore, is to effectively manage these inherent trade-offs. Our results demonstrate that while simple MTL methods struggle with this compromise, AIM's adaptive policy consistently pushes the performance frontier, outperforming other MTL heuristics in the data-scarce regimes most relevant to these challenges. AIM's contribution is therefore twofold. First, it offers superior performance within this necessary, yet inherently compromised, MTL paradigm. Second, and critically, its interpretable policy (as seen in Figures \ref{fig:qm9_heatmap} and \ref{fig:tpd_heatmap}) provides a diagnostic tool to understand the very task conflicts that necessitate this compromise in the first place. AIM offers not just a better answer, but deeper insight into the complex inter-task relationships that govern multi-property molecular design.

\subsection{Outlook and future directions}

The insights gained from this analysis directly inform the most promising future applications of AIM. The demonstrated ability of the policy to identify task groups (Figure \ref{fig:qm9_heatmap}, \ref{fig:tpd_heatmap}) provides a data-driven foundation for automated curriculum design. An early stage policy could be used to pre-train a model on synergistic task groups before introducing more conflicting objectives, a strategy that is more principled than manual curriculum design.

Similarly, the policy's function as a scientific hypothesis engine is grounded in its proven ability to learn known physical relationships. Its potential to uncover unexpected task compatibilities, by learning a permissive policy between two seemingly unrelated tasks, is therefore not mere speculation. It represents a concrete mechanism for generating machine-driven hypotheses that can guide targeted experimental validation, turning the optimizer into an active participant in the scientific discovery process. 

We also acknowledge several limitations that provide clear avenues for future research. A thorough ablation study is needed to isolate the specific contributions of the guidance loss and the regularizers within our augmented objective. Furthermore, a detailed analysis of the computational overhead introduced by the policy and its guidance set would be crucial for practical deployment. Finally, evaluating AIM's generalizability across diverse model architectures beyond the GNN used here will be essential to validate its broader utility as a model-agnostic optimizer.

\clearpage

\begin{ack}
We thank Andrea Anelli, Leonid Komissarov, Natasa Tagasovska, Pan Kessel, and Kenneth Atz for inspiring discussions. This study was financially supported by the Roche Access to Distinguished Scientists (ROADS) Program (grant no. ROADS-073 to G.S.).
\end{ack}

\bibliographystyle{unsrtnat}
\bibliography{references}

\begin{thebibliography}{15}
\providecommand{\natexlab}[1]{#1}
\providecommand{\url}[1]{\texttt{#1}}
\expandafter\ifx\csname urlstyle\endcsname\relax
  \providecommand{\doi}[1]{doi: #1}\else
  \providecommand{\doi}{doi: \begingroup \urlstyle{rm}\Url}\fi

\bibitem[Schneider et~al.(2020)Schneider, Walters, Plowright, Sieroka, Listgarten, Goodnow~Jr, Fisher, Jansen, Duca, Rush, et~al.]{schneider2020rethinking}
Petra Schneider, W~Patrick Walters, Alleyn~T Plowright, Norman Sieroka, Jennifer Listgarten, Robert~A Goodnow~Jr, Jasmin Fisher, Johanna~M Jansen, Jos{\'e}~S Duca, Thomas~S Rush, et~al.
\newblock Rethinking drug design in the artificial intelligence era.
\newblock \emph{Nature reviews drug discovery}, 19\penalty0 (5):\penalty0 353--364, 2020.

\bibitem[Atz et~al.(2024)Atz, Nippa, M{\"u}ller, Jost, Anelli, Reutlinger, Kramer, Martin, Grether, Schneider, et~al.]{atz2024geometric}
Kenneth Atz, David~F Nippa, Alex~T M{\"u}ller, Vera Jost, Andrea Anelli, Michael Reutlinger, Christian Kramer, Rainer~E Martin, Uwe Grether, Gisbert Schneider, et~al.
\newblock Geometric deep learning-guided suzuki reaction conditions assessment for applications in medicinal chemistry.
\newblock \emph{RSC Medicinal Chemistry}, 15\penalty0 (7):\penalty0 2310--2321, 2024.

\bibitem[Nippa et~al.(2025)Nippa, Atz, Stenzhorn, M{\"u}ller, Tosstorff, Benz, Binch, B{\"u}rkler, Haider, Heer, et~al.]{nippa2025expediting}
David~F Nippa, Kenneth Atz, Yannick Stenzhorn, Alex~T M{\"u}ller, Andreas Tosstorff, J{\"o}rg Benz, Hayley Binch, Markus B{\"u}rkler, Achi Haider, Dominik Heer, et~al.
\newblock Expediting hit-to-lead progression in drug discovery through reaction prediction and multi-objective molecular optimization.
\newblock 2025.

\bibitem[Anelli et~al.(2024)Anelli, Dietrich, Ectors, Stowasser, Bereau, Neumann, and van~den Ende]{anelli2024robust}
Andrea Anelli, Hanno Dietrich, Philipp Ectors, Frank Stowasser, Tristan Bereau, Marcus Neumann, and Joost van~den Ende.
\newblock Robust and efficient reranking in crystal structure prediction: a data driven method for real-life molecules.
\newblock \emph{CrystEngComm}, 26\penalty0 (41):\penalty0 5845--5849, 2024.

\bibitem[Minot and Reddy(2024)]{minot2024meta}
Mason Minot and Sai~T Reddy.
\newblock Meta learning addresses noisy and under-labeled data in machine learning-guided antibody engineering.
\newblock \emph{Cell systems}, 15\penalty0 (1):\penalty0 4--18, 2024.

\bibitem[Komissarov et~al.(2024)Komissarov, Manevski, Groebke~Zbinden, Schindler, Zitnik, and Sach-Peltason]{komissarov2024actionable}
Leonid Komissarov, Nenad Manevski, Katrin Groebke~Zbinden, Torsten Schindler, Marinka Zitnik, and Lisa Sach-Peltason.
\newblock Actionable predictions of human pharmacokinetics at the drug design stage.
\newblock \emph{Molecular Pharmaceutics}, 21\penalty0 (9):\penalty0 4356--4371, 2024.

\bibitem[Minot et~al.(2025)Minot, Stenzhorn, Wolfard, Strobel, Jablonski, Zimmerli, Binder, Grether, Martin, M{\"u}ller, et~al.]{minot2025active}
Mason Minot, Yannick Stenzhorn, Jens Wolfard, Sebastian Strobel, Philippe Jablonski, Daniel Zimmerli, Martin Binder, Uwe Grether, Rainer~E Martin, Alex~T M{\"u}ller, et~al.
\newblock Active and geometric deep learning advances chemical reaction prediction in data-scarce drug discovery.
\newblock 2025.

\bibitem[Allenspach et~al.(2024)Allenspach, Hiss, and Schneider]{allenspach2024neural}
Stephan Allenspach, Jan~A Hiss, and Gisbert Schneider.
\newblock Neural multi-task learning in drug design.
\newblock \emph{Nature Machine Intelligence}, 6\penalty0 (2):\penalty0 124--137, 2024.

\bibitem[Liu et~al.(2021)Liu, Liu, Jin, Stone, and Liu]{liu2021conflict}
Bo~Liu, Xingchao Liu, Xiaojie Jin, Peter Stone, and Qiang Liu.
\newblock Conflict-averse gradient descent for multi-task learning.
\newblock \emph{Advances in Neural Information Processing Systems}, 34:\penalty0 18878--18890, 2021.

\bibitem[Yu et~al.(2020)Yu, Kumar, Gupta, Levine, Hausman, and Finn]{yu2020gradient}
Tianhe Yu, Saurabh Kumar, Abhishek Gupta, Sergey Levine, Karol Hausman, and Chelsea Finn.
\newblock Gradient surgery for multi-task learning.
\newblock \emph{Advances in neural information processing systems}, 33:\penalty0 5824--5836, 2020.

\bibitem[Navon et~al.(2022)Navon, Shamsian, Achituve, Maron, Kawaguchi, Chechik, and Fetaya]{navon2022multi}
Aviv Navon, Aviv Shamsian, Idan Achituve, Haggai Maron, Kenji Kawaguchi, Gal Chechik, and Ethan Fetaya.
\newblock Multi-task learning as a bargaining game.
\newblock \emph{arXiv preprint arXiv:2202.01017}, 2022.

\bibitem[Ramakrishnan et~al.(2014)Ramakrishnan, Dral, Rupp, and Von~Lilienfeld]{ramakrishnan2014quantum}
Raghunathan Ramakrishnan, Pavlo~O Dral, Matthias Rupp, and O~Anatole Von~Lilienfeld.
\newblock Quantum chemistry structures and properties of 134 kilo molecules.
\newblock \emph{Scientific data}, 1\penalty0 (1):\penalty0 1--7, 2014.

\bibitem[Peteani et~al.(2024)Peteani, Huynh, Gerebtzoff, and Rodr{\'\i}guez-P{\'e}rez]{peteani2024application}
Giulia Peteani, Minh Tam~Davide Huynh, Gr{\'e}gori Gerebtzoff, and Raquel Rodr{\'\i}guez-P{\'e}rez.
\newblock Application of machine learning models for property prediction to targeted protein degraders.
\newblock \emph{Nature communications}, 15\penalty0 (1):\penalty0 5764, 2024.

\bibitem[Liu et~al.(2023)Liu, Feng, Stone, and Liu]{liu2023famo}
Bo~Liu, Yihao Feng, Peter Stone, and Qiang Liu.
\newblock Famo: Fast adaptive multitask optimization.
\newblock \emph{Advances in Neural Information Processing Systems}, 36:\penalty0 57226--57243, 2023.

\bibitem[Finn et~al.(2017)Finn, Abbeel, and Levine]{finn2017model}
Chelsea Finn, Pieter Abbeel, and Sergey Levine.
\newblock Model-agnostic meta-learning for fast adaptation of deep networks.
\newblock In \emph{International conference on machine learning}, pages 1126--1135. PMLR, 2017.

\end{thebibliography}

\clearpage 

\appendix

\section{Appendix}

\subsection{Algorithm}

The AIM update step is summarized in Algorithm 1.

\begin{algorithm}
\caption{AIM Update Step}
\label{alg:aim_update}
\begin{algorithmic}[1]
\Require Main model parameters $\theta$, Policy parameters $\Phi$ (containing $\tau$ or $\tau_{ij}$)
\Require Learning rates $\eta_\theta, \eta_\Phi$
\Require Hyperparameters $\lambda_g, \lambda_m, \lambda_p$
\Require Batch from primary training set $D_{\text{primary}}$, Batch from policy guidance set $D_{\text{guide}}$

\State // \textbf{Step 1: Compute Raw Task Gradients on Primary Data}
\For{each task $i \in \{1, \dots, N\}$}
    \State Compute task loss $\mathcal{L}_i(\theta)$ on $D_{\text{primary}}$
    \State Compute raw task gradient $\mathbf{g}_i = \nabla_\theta \mathcal{L}_i(\theta)$ \Comment{Stop gradient applied to $\mathbf{g}_i$ for policy input}
\EndFor

\State // \textbf{Step 2: Differentiable Gradient Intervention (Policy $\Psi$)}
\State Initialize $\mathbf{g}_{\text{intervened}} = \mathbf{0}$
\For{each task $i \in \{1, \dots, N\}$}
    \State Initialize $\mathbf{g}_i' = \mathbf{g}_i$
    \For{each conflicting task $j \in \{1, \dots, N\}, j \neq i$}
        \State Compute cosine similarity: $\cos(\mathbf{g}_i, \mathbf{g}_j) = \frac{\mathbf{g}_i \cdot \mathbf{g}_j}{\|\mathbf{g}_i\| \|\mathbf{g}_j\|}$
        \State Retrieve learnable threshold $\tau_{ij}$ (or global $\tau$ for scalar policy)
        \State Compute projection weight: $w_{\text{proj}}^{(i,j)} = \sigma\left(k \cdot (\tau_{ij} - \cos(\mathbf{g}_i, \mathbf{g}_j))\right)$
        \State Compute vector projection: $\text{proj}_{\mathbf{g}_j}(\mathbf{g}_i) = \left(\frac{\mathbf{g}_i \cdot \mathbf{g}_j}{\|\mathbf{g}_j\|^2}\right) \mathbf{g}_j$
        \State Modify gradient: $\mathbf{g}_i' \leftarrow \mathbf{g}_i' - w_{\text{proj}}^{(i,j)} \cdot \text{proj}_{\mathbf{g}_j}(\mathbf{g}_i)$
    \EndFor
    \State $\mathbf{g}_{\text{intervened}} \leftarrow \mathbf{g}_{\text{intervened}} + \mathbf{g}_i'$
\EndFor

\State // \textbf{Step 3: Compute Policy Loss on Guidance Data}
\State Compute task losses $\mathcal{L}_1(\theta), \dots, \mathcal{L}_N(\theta)$ on $D_{\text{guide}}$
\State // Policy Guidance Objective
\State $\mathcal{L}_{\text{guide}} = \sum_{i=1}^N \mathcal{L}_i(\theta)$
\State // Magnitude Preservation
\State $\mathcal{L}_{\text{magnitude}} = \left(\|\mathbf{g}_{\text{intervened}}\| - \sum_{i=1}^N \|\mathbf{g}_i\|\right)^2$
\State // Progress Penalty
\State Initialize $\mathcal{L}_{\text{progress}} = 0$
\For{each task $i \in \{1, \dots, N\}$}
    \State $\alpha_i = \mathcal{L}_i(\theta) / \sum_{j=1}^N \mathcal{L}_j(\theta)$
    \State $p_i = \mathbf{g}_{\text{intervened}} \cdot \mathbf{g}_i$
    \State $\mathcal{L}_{\text{progress}} \leftarrow \mathcal{L}_{\text{progress}} - \alpha_i \cdot p_i$
\EndFor
\State Total Policy Loss: $\mathcal{L}_{\text{policy}} = \lambda_g \mathcal{L}_{\text{guide}} + \lambda_m \mathcal{L}_{\text{magnitude}} + \lambda_p \mathcal{L}_{\text{progress}}$

\State // \textbf{Step 4: Update Parameters}
\State Update policy parameters: $\Phi \leftarrow \text{Optimizer}(\Phi, \nabla_\Phi \mathcal{L}_{\text{policy}}, \eta_\Phi)$ \Comment{e.g., Adam or SGD} 
\State Update main model parameters: $\theta \leftarrow \text{Optimizer}(\theta, \mathbf{g}_{\text{intervened}}, \eta_\theta)$ \Comment{e.g., Adam or SGD}

\end{algorithmic}
\end{algorithm}

\subsection{Training details and hyperparameters}

All models were trained on the ETH Z{\"u}rich Euler HPC using a single NVIDIA A100 GPU. The Adam optimizer was used for all experiments. The learning rate was managed by a ReduceLROnPlateau scheduler, which reduced the learning rate when the mean validation loss across all tasks plateaued. For the AIM policy optimizer, we used a CosineAnnealingWarmRestarts scheduler with the following parameters: $T_0=20$ epochs, $T_{mult}=1$, and $\eta_{min}=10^{-6}$.

Specific hyperparameters for the main model and the AIM policy were tuned for each dataset and are detailed in Table~\ref{tab:hyperparameters}.

\begin{table}[h!]
\centering
\caption{Hyperparameter settings.}
\label{tab:hyperparameters}
\begin{tabular}{lc}
\toprule
\textbf{Hyperparameter} & \textbf{Value} \\
\midrule
\multicolumn{2}{l}{\textit{General Training}} \\
Epochs & [300, 400] \\
Early Stopping Patience & 75 epochs \\
Main Model Learn Rate & [$1 \times 10^{-3}$, $5 \times 10^{-4}$] \\
\midrule
\multicolumn{2}{l}{\textit{AIM Policy Specific}} \\
Policy Learn Rate & [$5 \times 10^{-4}$, $1 \times 10^{-3}$, $5 \times 10^{-3}$] \\
$k$ (temperature) & 10 \\
$\lambda_g$ ($\mathcal{L}_{guide}$) & [0, 1] \\
$\lambda_m$ (magnitude preservation) & [0, 0.01] \\
$\lambda_p$ (progress penalty) & [0, 0.08, 0.8] \\
\bottomrule
\end{tabular}
\end{table}

\subsection{Additional extended results}

\begin{table*}[htbp] 
 \centering 
 \caption{Mean Absolute Error (MAE) on the QM9 test set. Models were trained on the 100k subset. Lower values are better.} 
 \label{tab:qm9_100k_tasks} 
 \setlength{\tabcolsep}{4pt} 
 \small 
 \vspace{5pt} 
 \begin{tabular}{lccccccccccc} 
 \toprule 
 Method & $\mu$ & $\alpha$ & $\epsilon_{HOMO}$ & $\epsilon_{LUMO}$ & $R^2$ & ZPVE & $U_0$ & $U$ & $H$ & $G$ & $C_v$ \\ 
 \midrule 
 STL & 0.067 & 0.181 & 60.58 & 53.91 & 0.503 & 4.54 & 58.84 & 64.24 & 63.85 & 66.22 & 0.072 \\ 
 \midrule 
 LS & 0.10 & 0.31 & 70.98 & 86.65 & 4.84 & 12.94 & 137.3 & 138.2 & 138.7 & 133.6 & 0.12 \\ 
 PCGrad & 0.11 & 0.31 & 78.72 & 91.31 & 3.98 & 8.97 & 121.1 & 121.8 & 122.0 & 119.0 & 0.11 \\ 
 NASH-MTL & 0.11 & 0.33 & 77.46 & 93.78 & 5.23 & 14.39 & 148.4 & 149.2 & 149.5 & 144.6 & 0.13 \\ 
 FAMO & 0.39 & 0.37 & 147.1 & 134.7 & \textbf{1.16} & \textbf{4.83} & \textbf{68.65} & \textbf{69.14} & \textbf{69.08} & \textbf{68.42} & 0.12 \\ 
 CAGrad & 0.11 & 0.32 & 79.82 & 94.89 & 3.34 & 7.34 & 116.2 & 116.8 & 117.1 & 114.1 & 0.12 \\ 
 \midrule 
 AIM (Scalar) & 0.089 & 0.268 & \textbf{59.37} & \textbf{71.29} & 4.18 & 9.82 & 111.5 & 111.6 & 112.0 & 113.2 & \textbf{0.103} \\ 
 AIM (Matrix) & \textbf{0.088} & \textbf{0.251} & 60.94 & 72.31 & 4.07 & 12.16 & 116.3 & 116.4 & 118.2 & 111.7 & \textbf{0.103} \\ 
 \bottomrule 
 \end{tabular} 
 \end{table*}

\begin{table*}[htbp]
\centering
\caption{Mean Absolute Error (MAE) for clearance tasks on the TPD ADME 100k subset. Lower values are better.}
\label{tab:tpd_100k_clearance}
\begin{tabular}{lcccccc}
\toprule
Method & rLM & hLM & mLM & pigLM & cyno & dLM \\
\midrule
STL & 0.098 & 0.105 & 0.101 & 0.077 & 0.096 & 0.088 \\
\hline
CAGrad & 0.123 $\pm$ 0.000 & 0.128 $\pm$ 0.002 & 0.124 $\pm$ 0.001 & 0.103 $\pm$ 0.002 & 0.118 $\pm$ 0.001 & 0.121 $\pm$ 0.001 \\
Nash-MTL & 0.123 $\pm$ 0.004 & 0.129 $\pm$ 0.004 & 0.124 $\pm$ 0.004 & 0.103 $\pm$ 0.004 & 0.119 $\pm$ 0.005 & 0.121 $\pm$ 0.004 \\
FAMO & 0.124 $\pm$ 0.002 & 0.133 $\pm$ 0.002 & 0.127 $\pm$ 0.002 & 0.104 $\pm$ 0.002 & 0.121 $\pm$ 0.003 & 0.123 $\pm$ 0.002 \\
LS & 0.126 $\pm$ 0.003 & 0.133 $\pm$ 0.005 & 0.127 $\pm$ 0.005 & 0.107 $\pm$ 0.004 & 0.122 $\pm$ 0.004 & 0.125 $\pm$ 0.005 \\
PCGrad & 0.137 $\pm$ 0.008 & 0.151 $\pm$ 0.010 & 0.136 $\pm$ 0.003 & 0.120 $\pm$ 0.009 & 0.129 $\pm$ 0.003 & 0.138 $\pm$ 0.006 \\
\midrule
AIM (Scalar) & 0.119 $\pm$ 0.003 & 0.125 $\pm$ 0.006 & \textbf{0.119 $\pm$ 0.001} & 0.101 $\pm$ 0.002 & 0.116 $\pm$ 0.004 & \textbf{0.118 $\pm$ 0.001} \\
AIM (Matrix) & \textbf{0.117 $\pm$ 0.003} & \textbf{0.123 $\pm$ 0.004} & 0.120 $\pm$ 0.003 & \textbf{0.100 $\pm$ 0.003} & \textbf{0.115 $\pm$ 0.004} & 0.119 $\pm$ 0.004 \\
\bottomrule
\end{tabular}
\end{table*}

\begin{table*}[htbp]
\centering
\caption{Mean Absolute Error (MAE) for distribution and binding tasks (Part 1) on the TPD ADME 100k subset. Lower values are better.}
\label{tab:tpd_100k_distribution_part1}
\begin{tabular}{lcccccc}
\toprule
Method & Fu-Rat & Fu-Hum & Fu-Mou & Fu-Dog & Fu-Mon \\
\midrule
STL & 0.083 & 0.083 & 0.080 & 0.081 & 0.081 \\
\hline
LS & 0.103 $\pm$ 0.005 & 0.102 $\pm$ 0.004 & 0.101 $\pm$ 0.005 & 0.101 $\pm$ 0.005 & 0.098 $\pm$ 0.005 \\
PCGrad & 0.113 $\pm$ 0.005 & 0.168 $\pm$ 0.064 & 0.148 $\pm$ 0.033 & 0.115 $\pm$ 0.008 & 0.148 $\pm$ 0.053 \\
Nash-MTL & 0.100 $\pm$ 0.003 & 0.099 $\pm$ 0.003 & 0.098 $\pm$ 0.003 & 0.097 $\pm$ 0.003 & 0.095 $\pm$ 0.004 \\
FAMO & \textbf{0.091 $\pm$ 0.002} & \textbf{0.090 $\pm$ 0.002} & \textbf{0.089 $\pm$ 0.002} & \textbf{0.089 $\pm$ 0.002} & \textbf{0.086 $\pm$ 0.002} \\
CAGrad & 0.102 $\pm$ 0.001 & 0.101 $\pm$ 0.001 & 0.100 $\pm$ 0.001 & 0.100 $\pm$ 0.001 & 0.097 $\pm$ 0.001 \\
\midrule
AIM (Scalar) & 0.097 $\pm$ 0.001 & 0.095 $\pm$ 0.001 & 0.095 $\pm$ 0.002 & 0.094 $\pm$ 0.001 & 0.092 $\pm$ 0.001 \\
AIM (Matrix) & 0.098 $\pm$ 0.002 & 0.096 $\pm$ 0.002 & 0.095 $\pm$ 0.002 & 0.095 $\pm$ 0.003 & 0.093 $\pm$ 0.002 \\
\bottomrule
\end{tabular}
\end{table*}

\begin{table*}[htbp]
\centering
\caption{Mean Absolute Error (MAE) for distribution and binding tasks (Part 2) on the TPD ADME 100k subset. Lower values are better.}
\label{tab:tpd_100k_distribution_part2}
\begin{tabular}{lcccccc}
\toprule
Method & Fu-HSA & Fu-brain & Fu-mic & LogP & LogD \\
\midrule
STL & 0.060 & 0.074 & 0.062 & 0.112 & 0.124 \\
\hline
LS & 0.088 $\pm$ 0.004 & 0.099 $\pm$ 0.004 & 0.083 $\pm$ 0.004 & 0.143 $\pm$ 0.006 & 0.156 $\pm$ 0.007 \\
PCGrad & 0.105 $\pm$ 0.005 & 0.110 $\pm$ 0.006 & 0.124 $\pm$ 0.041 & 0.177 $\pm$ 0.011 & 0.217 $\pm$ 0.057 \\
Nash-MTL & 0.086 $\pm$ 0.003 & 0.096 $\pm$ 0.004 & 0.081 $\pm$ 0.002 & \textbf{0.137 $\pm$ 0.005} & \textbf{0.150 $\pm$ 0.005} \\
FAMO & \textbf{0.077 $\pm$ 0.002} & \textbf{0.089 $\pm$ 0.002} & \textbf{0.072 $\pm$ 0.002} & 0.146 $\pm$ 0.003 & 0.161 $\pm$ 0.003 \\
CAGrad & 0.084 $\pm$ 0.000 & 0.097 $\pm$ 0.000 & 0.080 $\pm$ 0.000 & 0.140 $\pm$ 0.002 & 0.153 $\pm$ 0.002 \\
\midrule
AIM (Scalar) & 0.082 $\pm$ 0.001 & 0.094 $\pm$ 0.001 & 0.077 $\pm$ 0.001 & 0.140 $\pm$ 0.002 & 0.152 $\pm$ 0.001 \\
AIM (Matrix) & 0.083 $\pm$ 0.000 & 0.094 $\pm$ 0.003 & 0.077 $\pm$ 0.002 & 0.139 $\pm$ 0.004 & 0.154 $\pm$ 0.005 \\
\bottomrule
\end{tabular}
\end{table*}

\begin{table*}[htbp]
\centering
\caption{Mean Absolute Error (MAE) for permeability and metabolism tasks (Part 1) on the TPD ADME 100k subset. Lower values are better.}
\label{tab:tpd_100k_permeability_part1}
\begin{tabular}{lccccc}
\toprule
Method & MDCKv2 & MDCKv1 & Caco2 & MDR1 & PAMPA \\
\midrule
STL & 0.097 & 0.100 & 0.116 & 0.082 & 0.123 \\
\hline
LS & 0.123 $\pm$ 0.005 & 0.129 $\pm$ 0.004 & 0.153 $\pm$ 0.006 & 0.127 $\pm$ 0.007 & 0.155 $\pm$ 0.007 \\
PCGrad & 0.140 $\pm$ 0.006 & 0.193 $\pm$ 0.066 & 0.206 $\pm$ 0.005 & 0.142 $\pm$ 0.005 & 0.180 $\pm$ 0.010 \\
Nash-MTL & \textbf{0.120 $\pm$ 0.003} & \textbf{0.125 $\pm$ 0.003} & \textbf{0.148 $\pm$ 0.004} & 0.124 $\pm$ 0.005 & \textbf{0.149 $\pm$ 0.005} \\
FAMO & 0.130 $\pm$ 0.003 & 0.140 $\pm$ 0.004 & 0.171 $\pm$ 0.005 & 0.135 $\pm$ 0.005 & 0.175 $\pm$ 0.006 \\
CAGrad & \textbf{0.120 $\pm$ 0.001} & 0.126 $\pm$ 0.001 & 0.149 $\pm$ 0.001 & \textbf{0.123 $\pm$ 0.001} & 0.151 $\pm$ 0.002 \\
\midrule
AIM (Scalar) & 0.121 $\pm$ 0.000 & 0.128 $\pm$ 0.002 & 0.153 $\pm$ 0.002 & \textbf{0.123 $\pm$ 0.001} & 0.154 $\pm$ 0.003 \\
AIM (Matrix) & \textbf{0.120 $\pm$ 0.003} & 0.126 $\pm$ 0.002 & 0.151 $\pm$ 0.003 & \textbf{0.123 $\pm$ 0.002} & 0.151 $\pm$ 0.003 \\
\bottomrule
\end{tabular}
\end{table*}

\begin{table*}[htbp]
\centering
\caption{Mean Absolute Error (MAE) for permeability and metabolism tasks (Part 2) on the TPD ADME 100k subset. Lower values are better.}
\label{tab:tpd_100k_permeability_part2}
\begin{tabular}{lcccc}
\toprule
Method & kobs & CYP3A4 & CYP2C9 & CYP2D6 \\
\midrule
STL & 0.063 & 0.088 & 0.064 & 0.069 \\
\hline
LS & 0.108 $\pm$ 0.003 & 0.125 $\pm$ 0.008 & 0.095 $\pm$ 0.003 & 0.109 $\pm$ 0.003 \\
PCGrad & 0.104 $\pm$ 0.003 & 0.133 $\pm$ 0.006 & 0.172 $\pm$ 0.065 & 0.110 $\pm$ 0.006 \\
Nash-MTL & 0.107 $\pm$ 0.004 & 0.119 $\pm$ 0.006 & 0.093 $\pm$ 0.003 & 0.103 $\pm$ 0.002 \\
FAMO & 0.103 $\pm$ 0.002 & 0.119 $\pm$ 0.003 & 0.089 $\pm$ 0.001 & 0.100 $\pm$ 0.003 \\
CAGrad & \textbf{0.091 $\pm$ 0.002} & \textbf{0.107 $\pm$ 0.001} & \textbf{0.080 $\pm$ 0.002} & \textbf{0.094 $\pm$ 0.001} \\
\midrule
AIM (Scalar) & 0.101 $\pm$ 0.000 & 0.115 $\pm$ 0.004 & 0.089 $\pm$ 0.000 & 0.102 $\pm$ 0.003 \\
AIM (Matrix) & 0.101 $\pm$ 0.001 & 0.114 $\pm$ 0.001 & 0.090 $\pm$ 0.001 & 0.106 $\pm$ 0.007 \\
\bottomrule
\end{tabular}
\end{table*}


\clearpage

\end{document}